\documentclass{article}

\usepackage{microtype}
\usepackage{graphicx}
\usepackage{subcaption}
\usepackage{booktabs} 
\usepackage{csquotes}
\usepackage{multirow}

\usepackage{hyperref}

\usepackage[preprint]{icml2026}

\usepackage{amsmath}
\usepackage{amssymb}
\usepackage{mathtools}
\usepackage{amsthm}
\usepackage[capitalize,noabbrev]{cleveref}  

\theoremstyle{plain}

\theoremstyle{definition}

\theoremstyle{remark}

\icmltitlerunning{JEPA-Reasoner: Decoupling Latent Reasoning from Token Generation}

\begin{document}

\twocolumn[
  \icmltitle{JEPA-Reasoner: Decoupling Latent Reasoning from Token Generation}

  \icmlsetsymbol{equal}{*}

  \begin{icmlauthorlist}
    \icmlauthor{Bingyang Kelvin Liu}{equal,uiuc}
    \icmlauthor{Ziyu Patrick Chen}{equal,uiuc}
    \icmlauthor{David P. Woodruff}{equal,cmu}
  \end{icmlauthorlist}

  \icmlaffiliation{uiuc}{University of Illinois Urbana-Champaign, Urbana, USA}
  \icmlaffiliation{cmu}{Carnegie Mellon University, Pittsburgh, USA}

  \icmlcorrespondingauthor{David P. Woodruff}{dwoodruf@cs.cmu.edu}

  \icmlkeywords{Foundation Models, Latent Reasoning, ICML}

  \vskip 0.3in
]

\printAffiliationsAndNotice{}  

\begin{abstract}
Current autoregressive language models couple high-level reasoning and low-level token generation into a single sequential process, making the reasoning trajectory vulnerable to compounding expression errors. We propose JEPA-Reasoner, a novel architectural paradigm that decouples these tasks using a Joint-Embedding Predictive Architecture (JEPA) for pure latent-space reasoning and a separate Talker module for linguistic reconstruction. By isolating the reasoning engine from the discrete token-sampling process, our architecture enables: (1) Error Containment, where token-level failures cannot propagate into the latent reasoning chain; (2) Continuous Guidance, providing the generator with access to the entire lossless reasoning trajectory; and (3) Representation of Uncertainty, allowing the model to maintain multiple hypotheses via mixed latent vectors. Controlled experiments on synthetic and natural language tasks demonstrate that this decoupling enables a 0.9B model to achieve a 149.5\% improvement in 8-shot GSM8K accuracy over a coupled Transformer baseline trained on identical data. This work shifts the focus from scaling coupled models to investigating decoupled architectures as a more robust foundation for complex reasoning.
\end{abstract}

\section{Introduction}
The dominant paradigm for Large Language Models (LLMs) is a unified autoregressive process where internal reasoning and external expression are inextricably linked. In this coupled framework, the model must predict the next token based on a history that includes its own previously sampled discrete tokens. This creates a fundamental bottleneck: any localized error in token selection—whether a grammatical slip or a sub-optimal word choice—pollutes the context window, directly corrupting the trajectory of all subsequent reasoning.

While recent efforts such as COCONUT \cite{coconut} and Recurrent Depth models \cite{recurrent-latent-reasoning} have explored latent-space reasoning, they remain largely constrained by this coupling. They either interleave latent vectors with tokens in a single sequence or rely on causal masking that prevents early tokens from being informed by future reasoning states. Furthermore, these models struggle to represent uncertainty, as the discrete nature of token-level coupling often forces a collapse into a single reasoning path \cite{llms-reason-single-thread}.

In this paper, we propose a novel decoupled reasoning paradigm: JEPA-Reasoner, an architecture that fully decouples latent-space reasoning from token generation. Our approach utilizes the Joint-Embedding Predictive Architecture (JEPA) \cite{i-jepa} not merely for representation learning, but as an independent, autoregressive reasoning engine. The Reasoner operates entirely within a continuous, normalized latent space, generating a complete reasoning chain before any tokens are produced. A separate, lightweight Talker module then reconstructs human-readable text under the guidance of this completed latent trajectory.

By isolating these functions, we demonstrate three emergent properties of the decoupled architecture:
\begin{enumerate}
    \item \textbf{Mathematical Error Containment:} We provide a probabilistic factorization showing that because the reasoning chain is generated independently, token-sampling errors have no mathematical pathway to influence the logic of the Reasoner.
    \item \textbf{Architectural Efficiency:} Our experiments show that JEPA-Reasoner utilizes internally learned knowledge more effectively than coupled baselines. When trained on identical datasets, the decoupled architecture overcomes the "reasoning plateau" observed in small-scale models, showing superior in-context scaling from 5-shot to 8-shot tasks.
    \item \textbf{Mixed Latent States:} We empirically validate that the Reasoner can produce "mixed latent vectors" that encode multiple possible reasoning paths simultaneously, laying the foundation for multi-threaded reasoning without the overhead of beam search or multiple decodes.
\end{enumerate}

Rather than focusing on competitive SOTA benchmarks through massive scaling, this work serves as an architectural investigation into the benefits of decoupling. We show that even without sophisticated reinforcement learning or industrial training recipes, transitioning from a coupled to a decoupled paradigm yields significant gains in robustness and reasoning depth, suggesting a new path forward for foundation model design.

\section{Related Work}\label{sec:previous-work}
\paragraph{Joint-Embedding Predictive Architectures (JEPA).} JEPA \cite{i-jepa} introduced a framework that makes predictions in representation space. It utilizes self-supervised training to learn latent states that are not directly human-readable. A predictor module was trained to predict the target state based on encoded inputs. Multiple variants of this architecture have extended the JEPA family to various modalities and downstream tasks \cite{i-jepa}. However, these models are non-generative. Attempts to make JEPA generative, such as D-JEPA \cite{d-jepa}, use learned representations to condition diffusion models for data generation (e.g., text-to-image, text-to-audio), still fail to enable sequential reasoning or planning within the JEPA framework itself. In contrast, our key innovation is to adapt the core JEPA objective for autoregressive latent-space generation.

\paragraph{Latent Space Reasoning.} Previous work on latent space reasoning mainly focuses on looping hidden states, either through horizontal autoregression like COCONUT \cite{coconut}, or vertical recurrent depth scaling \cite{recurrent-latent-reasoning}. However, these paradigms utilize a single coupled model for both latent reasoning and token generation, ignoring the mismatch between the two tasks: one requires high-level global planning, decision-making, and choice tracing, while the other requires local correctness in grammar and fluency. Besides, whether iterating over tokens or deepening the computation per token, both paradigms remain bound by causal masking, meaning early token generation cannot be guided by future reasoning states. Given these limitations, our key innovation is to decouple latent-space reasoning and token generation. By generating a complete latent chain first, we enable consistent latent guidance, with all tokens generated from the whole reasoning trajectory. This produces a higher-quality answer that is less prone to error propagation. Additionally, the decoupled design enables efficient optimization with a single forward pass in latent space, unlike coupled models such as COCONUT or Recurrent Depth Transformers, which require multiple synchronized passes or complex recurrent unrolling.

\paragraph{Autoregressive Models and Robustness.} Modern Transformer models conduct token-level prediction in an autoregressive manner. While proven powerful on various tasks, this approach is known to suffer from compounding errors in long-horizon tasks. Techniques like Chain-of-Thought \cite{cot} improve reasoning by generating intermediate steps, but still operate at the token level. JEPA-Reasoner aims to enhance robustness of autoregressive generation by moving the reasoning process into a continuous, abstract latent space, reducing the impact of localized errors.

\section{Model Architecture}\label{sec:model-arch}
JEPA-Reasoner decouples the reasoning process from output generation, making next-state predictions completely dependent on previously generated, semantic-rich, lossless latent states. The architecture consists of:
\begin{itemize}
    \item \textbf{JEPA-Reasoner:} Generate sequential latent space reasoning chains independently.
    \item \textbf{Talker:} Translates the latent states into tokens. Note that the Talker can not make predictions. (Refer to ablation study in Appendix~\ref{app:indi-talker}) Its task is to reconstruct tokens solely based on the latent reasoning chain produced by JEPA-Reasoner. 
\end{itemize}

\subsection{JEPA-Reasoner}
\begin{figure*}[t]
    \centering
    \includegraphics[width=0.95\linewidth]{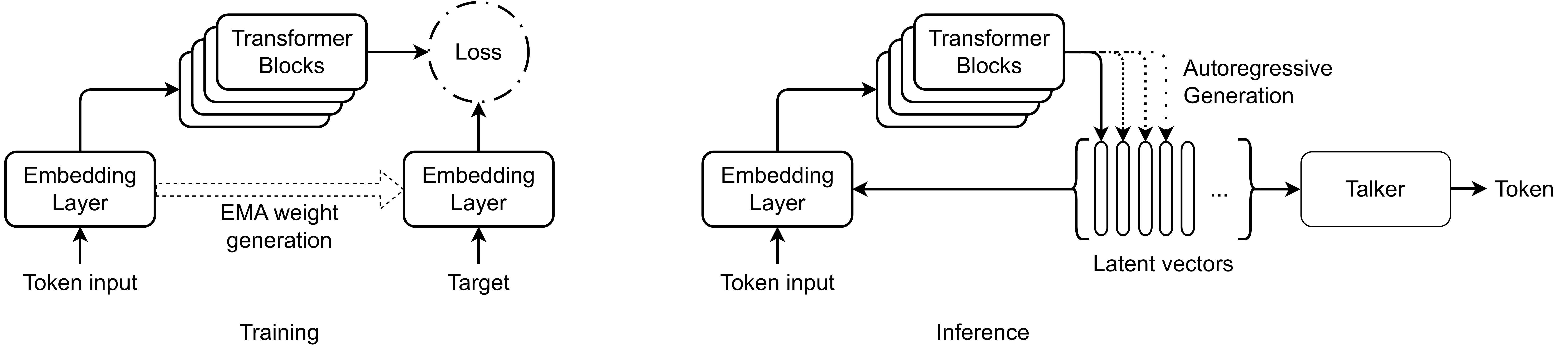}
    \caption{Architecture of JEPA-Reasoner and Talker. The Reasoner consists of an embedding layer as the token encoder and Transformer blocks as the predictor. The embedding layer for input tokens always uses the latest weights, while the weight of the embedding layer for target tokens is the exponential moving average of the historical weights of the input embedding layer.}
    \label{fig:reasoner-arch}
\end{figure*}

\paragraph{Model Components.} JEPA-Reasoner follows the JEPA philosophy, containing an embedding layer as a textual token encoder and modified Transformer blocks for the predictor, since the Transformer has proven its strong ability in sequence modeling. After the modified Transformer blocks, we applied a hybrid normalization layer (RMS and L2 normalization). We utilize L2 normalization to prevent exploding magnitude caused by residual connections. In the modified Transformer block, we apply a non-learnable QK-Norm \cite{qk-norm} to ensure numerical stability.

\paragraph{Latent Space Generation.} Unlike traditional JEPA models, in which the predictor is aimed at filling missing information in the current state \cite{i-jepa}, the predictor of JEPA-Reasoner generates the \emph{next} latent matrix representing the subsequent reasoning steps. Crucially, this generated latent matrix is not projected into vocabulary probabilities via an LM head. Instead, it is normalized by the hybrid normalization layer to the unit hypersphere and looped back as the input of the first Transformer block for the next round of autoregressive generation in the latent space.

\paragraph{Training Objective and Target Encoder.} The model is trained to predict latent representations provided by a target encoder. Following standard JEPA methodology, the target encoder weights are an exponential moving average (EMA) of the data encoder weights, providing stable and rich training targets. Given the normalized nature of our latent space, we use scaled cosine similarity loss computed entirely in the latent space, ensuring the predictor learns consistent feature representations and dynamics (refer to \cref{sec:sst} for more details).

\subsection{Talker Model}
The Talker model is a standard Transformer-based model trained independently. We designed two variants of Talker: Mono-Talker and Dual-Talker. Detailed information about the components of the two Talkers is shown in \cref{tab:talker-models}. Mono-Talker consists of decoders only, without an embedding layer or encoders. Mono-Talker is designed for reconstruction tasks that do not require context information, receiving latent vectors from JEPA-Reasoner and constructs the complete token sequence in one forward pass. Dual-Talker is designed for context-aware reconstruction, usually necessary in natural language tasks. It has an embedding layer, encoders, and decoders. The embedding layer is used for encoding previously determined outputs of JEPA-Reasoner that contain contextual information, while the encoder blocks receive latent vectors from JEPA-Reasoner as input. The decoders generate tokens autoregressively conditioned on previous tokens, with continuous latent guidance from the output of encoders. However, Dual-Talker was trained for reconstruction rather than generation, as our ablation study (Appendix~\ref{app:indi-talker}) showed that Talker is critically dependent on the Reasoner's output. During training, the JEPA-Reasoner is frozen. Talker receives the sequence of latent vectors and is trained to reconstruct the corresponding token sequence using standard cross-entropy loss.
\begin{table}[ht]
    \caption{Components of Mono-Talker and Dual-Talker}
    \label{tab:talker-models}
    \centering
    \begin{tabular}{lcc}
        \toprule
         & Mono-Talker & Dual-Talker \\
        \midrule
        Embedding Layer & No & Yes \\
        Standard Encoder & No & Yes \\
        Standard Decoder & Yes & Yes \\
        LM Head & Yes & Yes \\
        \bottomrule
    \end{tabular}
\end{table}

\section{Training Procedure}\label{sec:training}
The training process of JEPA-Reasoner consists of two main phases. (1) \textbf{Pretraining}, to acquire linguistic competence and world knowledge, and (2) \textbf{Self-Supervised Training (SST)}, to adapt the model for consistent reasoning in the continuous latent space.

\subsection{Pretraining}\label{sec:pretrain}
We apply established Transformer training methods to provide the model with basic knowledge and language understanding capabilities.

\paragraph{Objective and Methodology.} The model is trained as a standard decoder-only Transformer on the next-token prediction task via teacher forcing. To facilitate the transition to the subsequent SST phase, we introduce specific architectural changes:

\begin{itemize}
    \item \textbf{Tied Embeddings:} We share weights between the embedding layer and a temporary Language Model (LM) head. This LM head is discarded after pretraining.
    \item \textbf{L2 Normalization:} We disable the L2 normalization layer during this phase to maintain compatibility with standard Transformer training recipes.
\end{itemize}

The use of tied embeddings also indirectly facilitates angular alignment between the predicted vectors and the embedding vectors since it encourages $\mathbf{W}_{Embed}\cdot\mathbf{W}_{Embed}^T=I$ and $\mathbf{v}_{pred}\cdot\mathbf{v}_{embed} = \|\mathbf{v}_{pred}\|\cdot\|\mathbf{v}_{embed}\|\cdot\cos(\theta)$. This implicit angular alignment creates a smoother transition from token-level prediction to latent-level prediction in the subsequent SST phase.

\subsection{Self-Supervised Training}\label{sec:sst}

The SST phase transforms the model from a token generator into a pure latent-space reasoner. We adopt a self-supervised approach inspired by the JEPA series~\cite{i-jepa}. Unlike methods such as COCONUT~\cite{coconut}, our approach avoids the computational overhead of autoregressively generating final tokens or looping latent vectors multiple times for loss computation, enabling more efficient parallel training.

\paragraph{Optimization Objective.}
The model is now optimized to predict the latent representation of the next sequence segment. We discard the temporary LM head and restore the L2 normalization layer. We apply a similar self-supervised training strategy proposed in Meta's JEPA series \cite{i-jepa} and switch to scaled cosine distance loss, aligning with the L2 normalization strategy to ensure stability during autoregressive looping and encouraging the model to learn angular differences:

\begin{equation}
    \mathcal{L}(\theta, \theta') = k \cdot \left( 1 - \cos(h_{\text{pred}}(\theta), h_{\text{target}}(\theta')) \right)
\end{equation}

where $k$ is a scalar, $h_{\text{pred}}$ is the latent vector predicted by the Reasoner (parameters $\theta$) and $h_{\text{target}}$ is the target vector from the EMA encoder (parameters $\theta'$). We adopt a scaling factor of $k=4$ as we empirically observed that standard cosine distance yields insufficient gradients when the loss is small. (See Appendix~\ref{app:k-val} for detailed discussions about scaled cosine distance loss).

\paragraph{Target Generation (EMA).}
To prevent rank collapse while allowing angular adjustment, we employ an asymmetric weight update strategy:
\begin{enumerate}
    \item \textbf{Input Embedding (Online):} Updated via standard backpropagation.
    \item \textbf{Target Embedding (Target):} Updated via exponential moving average (EMA) of the input embedding weights. We applied a high momentum value of 0.98 to prevent rank collapse in the embedding layer while ensuring enough space to adjust for angular alignment.
\end{enumerate}

\section{Latent Space Properties}\label{sec:latent-property}
We analyze the property of JEPA-Reasoner's latent representation on two synthetic tasks designed to probe specific capabilities in controlled environments: mixed latent vector generation via a tree-search problem, and robustness to error propagation via a Context-Free Grammar (CFG) generation task.

\subsection{Continuous Representation of Uncertainty}
Within a reasoning process, JEPA-Reasoner is able to produce mixed latent vectors that are not limited to the discrete representations in the embedding layers. The mixed latent vectors approximate a linear combination of more than one vocabulary latent (latent vectors that correspond to individual vocabulary tokens). To systematically examine this behavior, we trained a smaller JEPA-Reasoner (42M) to search routes from the root to specific leaves in a binary tree.

\subsubsection{Data Preparation}
Training data consists of randomly generated binary trees with depth limited to 4. Each tree node was represented by a character with a unique token, making up the vocabulary along with other special tokens. In the generation process, we randomly pick node names to prevent the model from memorizing relationships based on names. Refer to Appendix~\ref{app:data-tree-search} for an example.

\subsubsection{Model Configuration}
The JEPA-Reasoner model and Mono-Talker model were built with specifications stated in \cref{tab:exp_tree_search_model_setup}. We chose the combination of JEPA-Reasoner with Mono-Talker because this task does not require context-aware reconstruction.
\begin{table*}[ht]
  \caption{Model configurations in tree-search experiment}
  \label{tab:exp_tree_search_model_setup}
  \centering
  \begin{tabular}{lccccc}
    \toprule
    ~ & Latent Dim. & Attention Dim. & FFN Dim. & Head Count & Decoder Count\\
    \midrule
    JEPA-Reasoner & 384 & 768 & 1536 & 16 & 18\\
    Mono-Talker & 384 & 768 & 1536 & 8 & 6\\
    \bottomrule
  \end{tabular}
\end{table*}

\subsubsection{Training}
The pretraining and SST process are completely the same as stated in \cref{sec:training}, except for loss masking. In the pretraining stage, loss was computed on all positions, while in SST, loss was only computed on the positions that define the desired route. When training the Talker model, we only passed latent vectors that describe the route to Talker to ensure it had no access to the tree structure or the target leaf, which guaranteed the Talker could not solve the task on its own.

\subsubsection{Results and Conclusions}
The final combination of the JEPA-Reasoner and Mono-Talker models achieved 99.87\% accuracy (exact match) in searching routes from the tree root to specific leaves. Given the restricted context window of the Mono-Talker model, we could confirm that only JEPA-Reasoner was responsible for reasoning. Based on this result, we examine the generated latent vectors to probe the reasoning behavior of JEPA-Reasoner.

\begin{figure}
    \centering
    \includegraphics[width=0.75\linewidth]{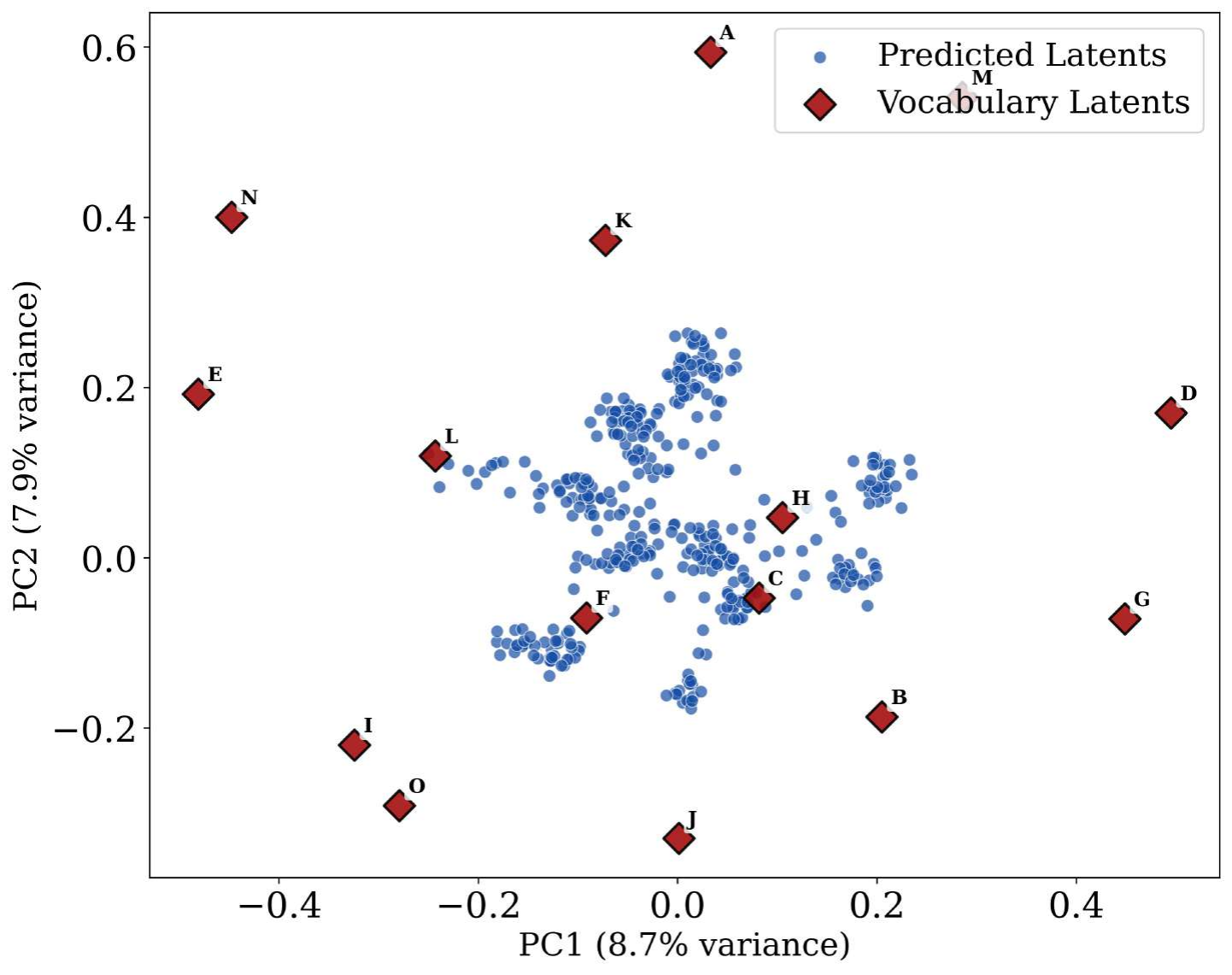}
    \caption{PCA analysis of latent representations of different tree leaves in the tree search experiment.}
    \label{fig:mix-latent-viz}
\end{figure}

\paragraph{Visualization of Mixed Latent Vectors} To visualize the mixed latent vectors, we extracted embedding vectors from distinct tree leaves alongside the model's output latent representations after one forward pass. Principal Component Analysis (PCA) was applied to the collected embeddings and model predictions, and \cref{fig:mix-latent-viz} demonstrates the visualization of first two principal components (PC1 and PC2). Noticing that predicted latent vectors (blue points in the figure) form a continuous cloud within the space spanned by discrete vocabulary embeddings (red diamond shapes). This distribution supports the experiment results that they are the linear combinations of vocabulary embeddings. Also, the predicted vectors do not converge to singular vocabulary points, providing empirical evidence for the hypothesis that JEPA-Reasoner is capable of maintaining information from multiple possible choices rather than committing to a single answer.

\paragraph{Quantitative Analysis} We calculated the distance from the predicted latent vector to the plane spanned by any two vocabulary vectors and sorted them from closest to farthest. In the sorted list, the plane spanned by latent vectors of sibling nodes frequently exhibits lower distances to the predicted latent vector, with an average ranking of top 1.72\% in the ordered list. Also, we figured out all coefficient sets, $\alpha$ and $\beta$, that satisfy $\alpha\cdot\mathbf{l}_0+\beta\cdot\mathbf{l}_1=\mathbf{l}_{proj}$, where $\mathbf{l}_0$ and $\mathbf{l}_0$ are latent vectors of sibling nodes and $\mathbf{l}_{proj}$ is the projection of the predicted latent vector on the spanned plane. After comparing $\alpha$ and $\beta$, we find that for 99.89\% of the time, the latent vector of the node on the correct route contributes more than the other sibling node. This discovery demonstrated that JEPA-Reasoner could make correct choices without completely discarding the other information that contains potentially correct choices. According to the previous COCONUT study \cite{coconut}, this behavior might lay the foundation for breadth-first multi-threaded reasoning.

\subsection{Robustness to Error Perturbation}\label{sec:robust-reasoner}
We demonstrate that decoupling reasoning generation from token production yields superior robustness. While coupled models must simultaneously maintain reasoning coherence and produce correct tokens, our decoupled approach allows the reasoning model to focus solely on consistency in the latent space, demonstrating superior generation quality and robustness to noisy input. We validate this advantage through two experiments targeting token-level errors and latent space noise, respectively.

\subsubsection{Experiment Methods}
\paragraph{Robustness Test for Token Level Error.}\label{sec:token-err-exp} To evaluate the robustness of the decoupled model on token-level errors in the input sequence, we randomly replace 0\% to 30\% ground truth tokens in the input sequence with incorrect tokens. We compare the JEPA-Reasoner against traditional Transformer models on multi-step completion tasks using the exact match metric. To minimize bias, we test on 5,248 samples randomly selected from a test set of 100,000.

\paragraph{Robustness Test for Latent Space Error.} To evaluate JEPA-Reasoner's robustness against perturbations in latent space, we compare its performance with the coupled continuous reasoning model COCONUT on multi-step completion tasks. In this setup, COCONUT generates four latent vectors followed by four tokens, whereas JEPA-Reasoner generates eight latent vectors and reconstructs eight tokens via the Talker module. We inject Gaussian noise at each step ($\mu=0$; $\sigma \in [0\%, 15\%]$ relative to the maximum value in each model's output latent vectors) and measure exact match accuracy on the final four tokens.

\subsubsection{Data Preparation}
Quantifying and testing token-level robustness in natural language is challenging because the impact of error tokens varies significantly by token type (e.g., replacing a keyword is more damaging than replacing a connector). To create a controlled experiment setup, we followed previous work \cite{allen-physics} and utilized Context-Free Grammar (CFG) for both experiments.

Our custom CFG features three terminal symbols with rule lengths of 3 or 4, producing complex and long sequences of 600--700 symbols. The diversity of sequences generated ensures that high accuracy requires learning the underlying structure rather than simple memorization (see Appendix~\ref{app:robustness-results} for detailed discussion).

\subsubsection{Model Configurations and Training Methods}
We define three model variants: the vanilla Transformer baseline ($T$), the coupled COCONUT model ($C$), and our decoupled JEPA-Reasoner ($R$). \cref{tab:cfg_model_config} shows detailed model configurations used in this section:

\begin{table}[ht]
    \caption{Model Configurations for CFG Task. Talker Block Count format (E+D) refers to Encoder and Decoder blocks in the Dual-Talker model.}
    \label{tab:cfg_model_config}
    \centering
    \begin{tabular}{lccc}
        \toprule
        & \textbf{$R$} & \textbf{$T$} & \textbf{$C$}\\
        \midrule
        Total Parameters & 315M & 338M & 338M \\
        \midrule
        Latent Dimension & 960 & 960 & 960 \\
        Attention Dimension & 960 & 960 & 960 \\
        FFN Dimension & 3840 & 3840 & 3840 \\
        Head Count & 16 & 16 & 16 \\
        \midrule
        Talker Block Count & 4 + 4 & -- & -- \\
        Reasoner Block Count & 16 & -- & -- \\
        Transformer Block Count & -- & 24 & 24 \\
        Total Blocks & 24 & 24 & 24 \\
        \bottomrule
    \end{tabular}
\end{table}

We first pretrain a Transformer on CFG data using cross-entropy loss. The checkpoint with the highest accuracy is selected as the initialization for all subsequent models \footnote{The JEPA-Reasoner is initialized using the first $N$ blocks of the pretrained Transformer to match its layer count.}. We then post-train $T$ as a baseline. $C$ is trained to predict a sequence of hidden states followed by tokens. $R$ is trained following the methodology in \cref{sec:training}. All training phases use identical hyperparameters (learning rate $1\times10^{-4}$, batch size 128, and context length 1024) and all models are trained until the loss stabilizes. Then the best-performing checkpoint is selected as the representative for each architecture.

\subsubsection{Results and Conclusions}
\begin{figure*}[t]
    \centering
    \includegraphics[width=0.95\linewidth]{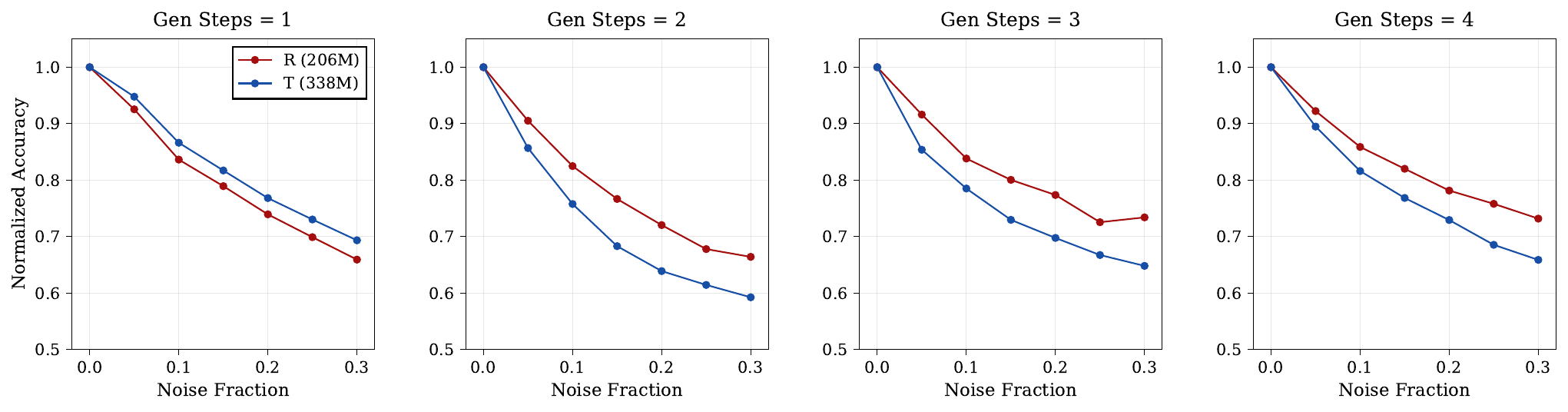}
    \caption{The relative performance of JEPA-Reasoner $R$ and Transformer baseline $T$ at each generation step with different fractions of correct tokens in the input sequence being replaced with wrong ones.}
    \label{fig:token-err-results}
\end{figure*}

The results of the above experiments confirm the robustness advantage of the decoupled architecture. In the token-level error experiment, JEPA-Reasoner demonstrated less performance degradation than the baseline ($T$) when facing input noise during multi-step CFG completion tasks (\cref{fig:token-err-results}). JEPA-Reasoner also exhibits higher performance across different magnitudes of Gaussian noise in the latent space error experiment (\cref{tab:latent-noise}), providing more empirical evidence for its robustness advantage.

\begin{table}[ht]
    \caption{Exact match accuracy of JEPA-Reasoner $R$ and COCONUT $C$ under different latent noise levels.}
    \label{tab:latent-noise}
    \centering
    \begin{tabular}{lcc}
        \toprule
         & $R$ & $C$ \\
         \midrule
        $\sigma=0.0$ & 0.4588 & 0.3740 \\
        $\sigma=0.05\times \max{(h_{t})}$ & 0.4681 & 0.3688 \\
        $\sigma=0.10\times \max{(h_{t})}$ & 0.4643 & 0.3650 \\
        $\sigma=0.15\times \max{(h_{t})}$ & 0.4468 & 0.3629 \\
        \bottomrule
    \end{tabular}
\end{table}

These results demonstrate that JEPA-Reasoner has the potential to address the limitations of existing paradigms discussed in \cref{sec:previous-work}. By operating in a normalized latent space and offloading token generation to the Talker module, subsequent reasoning outputs do not condition on previous decisions, thereby mitigating error accumulation in the autoregressive process and enabling robust sequential generation under noisy conditions.

\section{Real World Evaluation}
We evaluate the 694M JEPA-Reasoner and 198M Mono-Talker (0.9B combined) on natural language tasks following the training methods in \cref{sec:training}. Our analysis focuses on two key comparisons:
\begin{enumerate}
    \item \textbf{Comparison with Token-Level Model:} The performance difference between the fine-tuned Transformer model (token level prediction) and JEPA-Reasoner paired with its Mono-Talker model (latent level prediction).
    \item \textbf{Comparison with Other Reasoning Paradigms:} The performance of JEPA-Reasoner relative to other latent and non-latent reasoning models of similar scale.
\end{enumerate}

\subsection{Model Training}
We first pretrain a standard 0.9B Transformer model on C4 and Wikitext \cite{wikitext-dataset} for 300k steps, then fine-tune it for 42k steps using general math QA pairs. The best-performing checkpoint is selected to initialize and further fine-tune a Transformer model to obtain the baseline for comparison. We also trained a JEPA-Reasoner + Talker pair following the methodology in \cref{sec:training} with the same data and hyperparameters used to obtain the Transformer baseline. The best-performing checkpoints of each architecture that emerged during each training phase are selected as their representative.

\subsection{Performance Gain from Decoupled Architecture}
We evaluated both the fine-tuned Transformer model and JEPA-Reasoner on the GSM8K \cite{gsm8k} benchmark. \cref{tab:gsm8k-self-compare} shows the performance under both 5-shot and 8-shot settings.

\begin{table}[ht]
    \caption{GSM8K performance of Transformer baseline and JEPA-Reasoner}
    \label{tab:gsm8k-self-compare}
    \centering
    \begin{tabular}{lcc}
         \toprule
         \multirow{2}{*}{Model} & \multicolumn{2}{c}{Accuracy (\%)} \\
         \cmidrule(lr){2-3}
          & 5-shot & 8-shot \\
         \midrule
         Transformer Baseline & 20.7 & 20.8\\
         JEPA-Reasoner + Talker & 39.4 & 51.9\\
         \bottomrule
    \end{tabular}
\end{table}

\paragraph{Accuracy Gain.}\label{sec:accu-gain} JEPA-Reasoner substantially outperforms the fine-tuned Transformer, achieving performance gains of 79.2\% (5-shot) and 149.5\% (8-shot) on the GSM8K benchmark compared with the baseline. These results confirm that adapting to a decoupled, latent-space reasoning paradigm significantly improves the model's ability to utilize internally learned knowledge. 

Notably, both the Transformer baseline and JEPA-Reasoner are trained with identical data and hyperparameters to ensure fair comparison. Furthermore, the scaled cosine similarity loss used in SST prioritizes smooth latent transitions over logical correctness. Together, these factors suggest that the performance gain stems from the advantage of the decoupled latent reasoning architecture rather than new knowledge obtained during SST.

\paragraph{In Context Learning.} Crucially, the JEPA-Reasoner exhibits superior scaling behavior. While the Base Transformer's performance stagnates between 5-shot and 8-shot settings (improving only by 0.1\%), the JEPA-Reasoner utilizes the additional examples to achieve a 131.7\% improvement in performance. This scaling behavior indicates that the decoupled latent space reasoning architecture effectively overcomes the reasoning plateau, which was widely observed in small-scale token-based models, allowing for more robust logical deduction that is less constrained by surface-level token statistics.

\subsection{Performance Comparison with Other Reasoning Paradigms}
We evaluate JEPA-Reasoner against established reasoning paradigms like Chain-of-Thought (CoT) and Recurrent Depth. \cref{tab:gsm8k-vs-others} shows the GSM8K benchmark performance of our JEPA-Reasoner model (paired with Talker) relative to other models of comparable size or performance\footnote{Scores for Llama 3.2-1B \cite{llama3-2-1b-gsm8k} and Huginn-0125 \cite{recurrent-latent-reasoning} are cited from their official reports. As official 8-shot scores are unavailable for Qwen 3 0.6B, we evaluated it using \texttt{lm-evaluation-harness} \cite{lm-eval-harness}, employing the same configuration used for JEPA-Reasoner.}.

Comparison of JEPA-Reasoner against other reasoning paradigms on the GSM8K benchmark. Note that the COCONUT model is excluded as public 8-shot results are unavailable, and prior work indicates it trades performance for efficiency relative to CoT \cite{coconut}.

\begin{table*}[ht]
    \caption{Comparison of JEPA-Reasoner against other reasoning paradigms on the GSM8K benchmark. Note that the COCONUT model is excluded as neither its checkpoints nor its 8-shot GSM8K benchmark result are publicly available, and prior work indicates it trades performance for efficiency relative to CoT \cite{coconut}.}
    \label{tab:gsm8k-vs-others}
    \centering
    \begin{tabular}{llcc}
        \toprule
        Reasoning Paradigm & Example & Model Size & 8-shot Accuracy (\%) \\
        \midrule
        CoT & Llama 3.2 & 1B & 44.4 \\
        CoT & Qwen 3 & 0.6B & 42.5 \\
        Recurrent Depth & Huginn-0125 & 3.5B & 42.1 \\
        \textbf{Decoupled Latent Reasoning} & \textbf{JEPA-Reasoner (Ours)} & \textbf{0.9B} & \textbf{51.9} \\
        \bottomrule
    \end{tabular}
\end{table*}

\paragraph{Analysis of Results.} As illustrated in \cref{tab:gsm8k-vs-others}, JEPA-Reasoner demonstrates superior performance on the GSM8K benchmark compared to both CoT models and other latent reasoning architectures. With a parameter count of only 0.9B, our model achieves an 8-shot score of 51.9\%, outperforming the CoT baseline of the same size (Llama 3.2-1B) by 7.5 percentage points. While the Recurrent Depth model (Huginn-0125) offers a strong baseline at 42.1\%, it requires nearly four times the parameter count to achieve results that are still 9.8 percentage points lower than JEPA-Reasoner. Consequently, these results serve as strong empirical evidence that JEPA-Reasoner's decoupled latent reasoning architecture is capable of handling complex natural language reasoning tasks both  effectively and efficiently.

\section{Theoretical Advantage of Decoupled Architecture}
the robustness advantage of JEPA-Reasoner is built on two complementary mechanisms: the decoupled architecture design and the normalized property of latent space. We can formalize this advantage by analyzing the probabilistic assumptions and information flow within the coupled and decoupled paradigms. Denote $R = (r_1, r_2, \dots, r_T)$ as the sequence of latent reasoning states and $X = (\hat{x}_1, \hat{x}_2, \dots, \hat{x}_T)$ as the sequence of generated tokens.

\subsection{Decoupled Architecture Stops the Propagation of Error}
Assume that the joint probability can be factored into $P(R,X) = P(R) \cdot P(X | R)$. The autoregressive update rules for the reasoning state $r_t$ and token $\hat{x}_t$ at step $t$ are defined as:

$$
\begin{cases}
r_t = f_{\theta}(r_{t-1}) & \text{Reasoner Update} \\
\hat{x}_{t} \sim g_{\phi}(r_{t}, \hat{x}_{1:t-1}) & \text{Talker Sampling}
\end{cases}
$$

where $f_\theta$ is the deterministic predictor function of the Reasoner and $g_\phi$ is the probabilistic decoding function of the Talker. We define the sensitivity of the future reasoning state $r_{t+1}$ with respect to the current generated token $\hat{x}_t$ using partial derivatives as follows. Noticing that by the definition of decoupled architecture, $r_{t}$ depends solely on $r_{t-1}$. Considering that the generated token $\hat{x}_{t}$ does not appear in the input argument of $f_{\theta}$. Which means that $\frac{\partial r_{t}}{\partial \hat{x}_{\tau}} = 0 \quad \forall \tau \ge 1$. Using the chain rule, we obtain:

$$
\frac{\partial r_{t+1}}{\partial \hat{x}_t} = \frac{\partial f_\theta(r_t)}{\partial r_t} \cdot \underbrace{\frac{\partial r_t}{\partial \hat{x}_t}}_{0} = \mathbf{0}
$$

As a conclusion, there is no mathematical pathway for token-space errors produced during sampling to propagate into the latent reasoning trajectory, which enables better robustness during generation

\subsection{Bounded Error via Normalization and Angular Alignment}
In this section, we demonstrate that unlike standard autoregressive models where errors can compound in both magnitude and direction, JEPA-Reasoner’s error bound is explicitly enforced by the L2 normalization layer and implicitly minimized by the angular training objective. This dual constraint prevents the reasoning trajectory produced from diverging unbounded, providing the theoretical basis for the superior robustness observed in our experiments.

\paragraph{Explicitly Constrained by Architecture} By constraining all latent states to the unit hypersphere $\mathcal{S}^{d-1}$, the divergence between an arbitrary trajectory produced by JEPA-Reasoner during runtime (denoted by $R$) and an optimal reference trajectory (denoted by $R^*$) is strictly bounded. Since the predictor output is passed through an L2 normalization layer before looping back, every reasoning state $r_t$ satisfies $\|r_t\|_2 = 1$. Consequently, the Euclidean distance between the predicted state and the optimal state is physically constrained by the diameter of the hypersphere:

$$
\|r_{t} - r_{t}'\|_{2} \leq \text{diam}(S^{d-1}) = 2
$$

\paragraph{Implicitly Minimized by Training Objective} Furthermore, the scaled cosine distance loss $\mathcal{L}_t \propto 1 - \cos(r_t, r^*_t) = \frac{1}{2} \|r_t - r^*_t\|_2^2$ used in our Self-Supervised Training (SST) phase encourages the model to be aligned with a continuous trajectory on a bounded hypersphere instead of making leaps between discrete tokens like a traditional Transformer, implicitly reducing the risk of producing arbitrary latent vectors that diverges from the optimal reasoning path.

\section{Summary}
We introduce JEPA-Reasoner, a novel architecture that decouples latent space reasoning from token generation. Our approach enables continuous latent reasoning guidance while mitigating step-by-step error propagation. Efficient parallel training was also made possible without sacrificing latent reasoning performance compared with COCONUT. Our experiments suggest that by decoupling the high-level latent space reasoning process from low-level token generation, JEPA-Reasoner produced promising potential for multi-threaded reasoning and exhibited enhanced robustness to input noise and error accumulation when generating structured sequences.

\section*{Impact Statement}
This paper presents work whose goal is to advance the field of Machine Learning. There are many potential societal consequences of our work, none of which we feel must be specifically highlighted here.

\bibliography{citations}
\bibliographystyle{icml2026}

\newpage
\appendix
\onecolumn

\section{Data for Tree Search Experiment}\label{app:data-tree-search}
\begin{figure}
    \centering
    \includegraphics[width=0.65\linewidth]{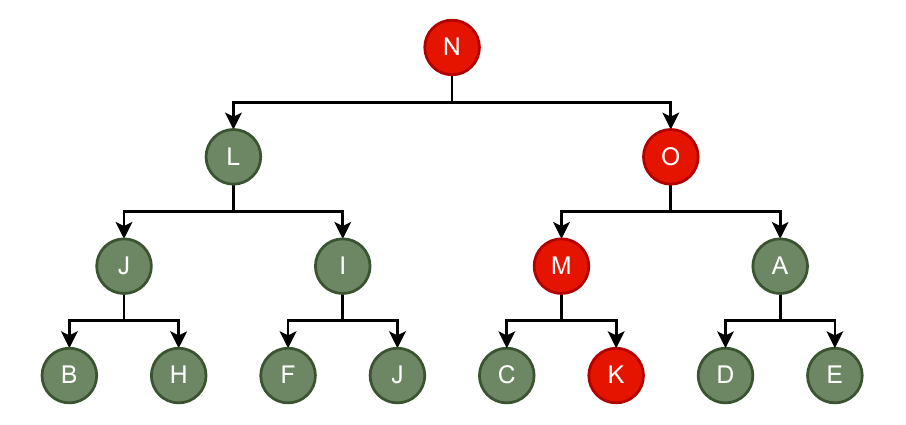}
    
    \caption{Visualization of the tree structure in the given example.}
    \label{fig:tree-viz}
    
\end{figure}

The following is an example of data used in the tree-search experiment:
\begin{center}
\texttt{NL,NO,LJ,LI,OA,OM,JB,JH,IG,IF,AD,A\\E,MK,MC[ROOT]N[TARGET]K[ROUTE]NOMK}
\end{center}
Visualization of the example can be seen in \cref{fig:tree-viz}. In the sequence, each character pair represents a parent-child node pair, with the former one being the parent node and the later one being the child. All pairs are separated by a comma. The searching task is specified after the tree-structure definition, with special token \texttt{[ROOT]} indicating the tree root, \texttt{[TARGET]} indicating which leaf to search for, and \texttt{[ROUTE]} stating the correct searching route. All characters, comma, \texttt{[ROOT]}, \texttt{[TARGET]}, and \texttt{[ROUTE]} have a corresponding token, making up the whole vocabulary for the model along with the padding token and the end-of-sentence token.

\section{Further Details for CFG Experiments}\label{app:robustness-results}
\subsection{CFG Rules and Sample}
\begin{figure}[ht]
    \begin{center}
    \includegraphics[width=0.65\textwidth]{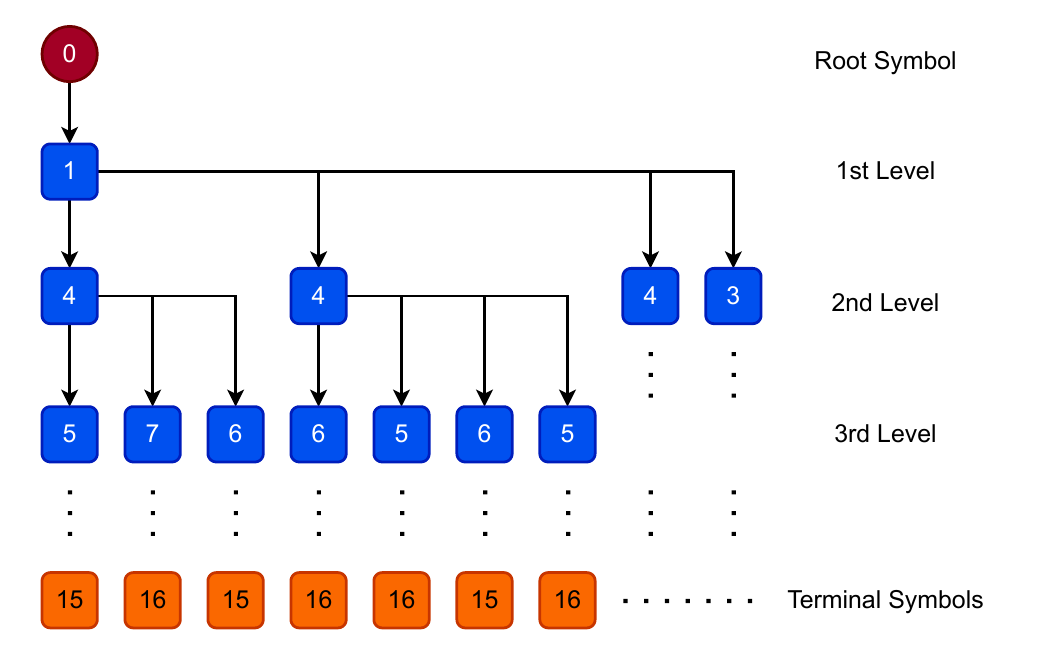}
    \end{center}
    
    \caption{A picture demonstrating how CFG sequence is generated. It involves replacing non-terminal symbols at each level with symbols from the next level according to a given rule.}
    \label{fig:cfg-sample}
    
\end{figure}

CFG can hierarchically produce highly structured expressions by replacing non-terminal symbols at each level with next-level symbols following a production rule, as shown in \cref{fig:cfg-sample}. A sequence of terminal symbols is considered to be valid if it can be transformed back to the root symbol with dynamic programming and the given production rule. The recursive structure and local ambiguity of CFG sequences enable them to model the rich and recursive structure in languages, including grammar and logic. We designed our own CFG following the method used by \cite{allen-physics}. The production rule used in our experiments is a five-level CFG production rule set featuring three terminal symbols with 3 or 4 rule lengths at each level, which typically generates long (typically 600 to 700 symbols per sample) and locally ambiguous sequences.

Since even a 5-level CFG production rule that allows each non-terminal symbol to produce 2 to 3 symbols in the next level (simpler than our 5-level production rule that allows each non-terminal symbol to produce 3 to 4 symbols in the next level) is capable of producing more than $4\times10^8$ distinctive sequences, we conclude that the models in the CFG experiments does not rely on memorizing possible sequences during training to achieve high accuracy on completion tasks.

Previous research \cite{allen-physics} shows that Transformer blocks can encode the structure of CFG rules within parameters. We assume that a robust model should be able to recognize the high-level structure of the input sequence, thus ignoring faulty tokens in the input. Since each high-level element in our CFG sequence produces 3 to 4 tokens, the model should be able to maintain relatively stable performance across at least 4 generation steps.

\subsection{A Sample CFG Sequence}
We demonstrate a sample CFG sequence from the training dataset: \texttt{15 16 16 14......14 16 14 (666 terminal symbols in total)}. It consists of three kinds of terminal symbols.

\subsection{Detailed Results of Token Level Error Test}
We provide the accuracy of Transformer baseline and JEPA-Reasoner under different noise levels on CFG completion task mentioned in \cref{sec:robust-reasoner} here in \cref{tab:token-level-err-raw-results}. Note that the accuracy is the total correct symbols generated divided by the total symbols needed in the generation task.

\begin{table*}[t]
    
    \caption{Absolute and relative (in parentheses, normalized to zero noise)  accuracy of Transformer baseline and JEPA-Reasoner under different noise levels on CFG completion task}
    \label{tab:token-level-err-raw-results}
    
    \centering
    \begin{tabular}{cc|cccc}
        \toprule
        \multirow{2}{*}{\textbf{Noise}} & \multirow{2}{*}{\textbf{Model}} & \textbf{Gen=1} & \textbf{Step=2} & \textbf{Step=3} & \textbf{Step=4} \\
        \cmidrule(lr){3-6}
        0.00 & $T$ & 0.9230 (1.00) & 0.9181 (1.00) & 0.6185 (1.00) & 0.6861 (1.00) \\
        & $R$ & 0.9191 (1.00) & 0.7405 (1.00) & 0.5722 (1.00) & 0.6646 (1.00) \\
        \cmidrule(lr){2-6}
        0.05 & $T$ & 0.8748 (0.95) & 0.7867 (0.86) & 0.5281 (0.85) & 0.6140 (0.89) \\
        & $R$ & 0.8506 (0.93) & 0.6703 (0.91) & 0.5242 (0.92) & 0.6130 (0.92) \\
        \cmidrule(lr){2-6}
        0.10 & $T$ & 0.7996 (0.87) & 0.6958 (0.76) & 0.4857 (0.79) & 0.5600 (0.82) \\
         & $R$ & 0.7688 (0.84) & 0.6109 (0.82) & 0.4797 (0.84) & 0.5708 (0.86) \\
        \cmidrule(lr){2-6}
        0.15 & $T$ & 0.7543 (0.82) & 0.6273 (0.68) & 0.4514 (0.73) & 0.5273 (0.77) \\
        & $R$ & 0.7256 (0.79) & 0.5678 (0.77) & 0.4581 (0.80) & 0.5451 (0.82) \\
        \cmidrule(lr){2-6}
        0.20 & $T$ & 0.7092 (0.77) & 0.5868 (0.64) & 0.4316 (0.70) & 0.5005 (0.73) \\
        & $R$ & 0.6797 (0.74) & 0.5336 (0.72) & 0.4428 (0.77) & 0.5195 (0.78) \\
        \cmidrule(lr){2-6}
        0.25 & $T$ & 0.6742 (0.73) & 0.5643 (0.61) & 0.4129 (0.67) & 0.4704 (0.69) \\
        & $R$ & 0.6426 (0.70) & 0.5021 (0.68) & 0.4152 (0.73) & 0.5039 (0.76) \\
        \cmidrule(lr){2-6}
        0.30 & $T$ & 0.6402 (0.69) & 0.5442 (0.59) & 0.4010 (0.65) & 0.4521 (0.66) \\
        & $R$ & 0.6061 (0.66) & 0.4919 (0.66) & 0.4201 (0.73) & 0.4865 (0.73) \\
        \bottomrule
    \end{tabular}
\end{table*}

\section{Ablation Study of Talker Model}\label{app:indi-talker}
We conduct an ablation study to verify two critical properties of our decoupled architecture: \begin{enumerate}
\item \textbf{Reasoning Dominance:} The reasoning process is strictly driven by the JEPA-Reasoner's latent trajectory, preventing the Talker from bypassing the architecture to perform independent inference.
\item \textbf{Linguistic Capability:} Talker module acts as an effective \enquote{Language Interface,} capable of translating abstract latent vectors into grammatically correct and semantically coherent natural language.
\end{enumerate}

We test this by corrupting the output of Reasoner in different ways. Our empirical experiments show strong evidence that the Talker cannot reason on its own and serves primarily as a readout mechanism for the Reasoner's planning.

\subsection{Experiment Setup}
Using the training method mentioned in \cref{sec:training}, we use the JEPA-Reasoner model initialized with Transformer blocks trained on C4 and Wikitext \cite{wikitext-dataset} dataset in this experiment setup to produce a human-readable result. We conducted controlled experiments using two sample inputs from the training dataset to evaluate the dependency of the Talker model on the Reasoner's output: 

The first sample is: \enquote{Francis Bacon was an English philosopher and statesman who served as Attorney General and Lord Chancellor of England under King James I. Bacon argued for the importance of natural philosophy, guided by the scientific, his works remained influential}. This sample is used as JEPA-Reasoner and Talker's input unless mentioned otherwise. The second sample is \enquote{Jean-Paul Sartre was a French philosopher, political activist, biographer, and literary critic. Sartre was one of the key figures in the philosophy of existentialism (and phenomenology).}, which is used in the \enquote{Semantic Mismatch} experiment

We systematically corrupted different components of the input to isolate the contribution of each part:
\begin{itemize}
    \item Baseline: Normal operation with clean Reasoner output
    \item Random String Replacement: Replace Reasoner output with a random string
    \item Initial Token Corruption: Keep Reasoner output clean, but replace Talker's initial input string (decoder input) with a random string.
    \item Gaussian Noise: Replace Reasoner output with Gaussian noise ($\mu=0$, $\sigma=1$)
    \item Semantic Mismatch: Use Reasoner output from a different sentence
\end{itemize}

\subsection{Results}
\cref{tab:talker-ablation} presents the results of our ablation experiments. The Talker's initial input tokens were the first 10 tokens of the Francis Bacon sample (\textit{Francis Bacon was an English philosopher and statesman who} in natural language across most experiments unless mentioned otherwise.

\begin{table}[ht]
    
    \caption{Ablation Study Results: Talker Model Output Under Different Input Corruptions.}
    \label{tab:talker-ablation}
    
    \centering
        \begin{tabular}{lp{8cm}}
            \toprule
            Condition & Talker Output \\
            \midrule
            Baseline & Bacon was an English philosopher and statesman who served as Attorney General and Lord Chancellor of England under King James I. Bacon argued for the importance of natural philosophy, guided by the scientific; his works remained influential \\
            \midrule
            Random String & Francis Bacon was an English philosopher and statesman whoTGvIUujapaUYDIUbIyuTviuiYtvU \\
            \midrule
            Random Initial Tokens & SsDXicoundfdfiyx served as Attorney General and Lord Chancellor of England under King James I. Richardson argued for the importance of natural philosophy, guided by the scientific, his works remained influential \\
            \midrule
            Gaussian Noise & Francis Bacon was an English philosopher and statesman who 0 not deathsardumn technicalipt'tputalase2006 (ightsringst 0q virtually lowesteral 1 new date 2007 0 results; \\
            \midrule
            Semantic Mismatch & Francis Bacon was an English philosopher and statesman who, political activist, biographer, and literary critic. Zartre was one of the key figures in the philosophy of existentialism (and phenomenology). \\
            \bottomrule
        \end{tabular}
    
\end{table}

\paragraph{Latent Dependency (Proof of No Independent Reasoning).} When the Reasoner output is replaced with Gaussian noise, the Talker produces incoherent output. This demonstrates that the Talker cannot generate meaningful content without proper latent representations.

\paragraph{Robustness to Token Errors (Proof of Linguistic Capability).} When only the initial input tokens are corrupted, the Talker can still produce largely coherent content, guided by the clean Reasoner output, although some localized errors occur (e.g., Richardson appears). This suggests the Reasoner's latent representations carry the primary semantic information.

\paragraph{Semantic Fidelity (The Decoupling Verification).} When using the Reasoner output from the Jean-Paul Sartre sample as Dual-Talker's input latent. Although the first 10 tokens from the Francis Bacon sample are used as the initial input tokens of decoder blocks in Dual-Talker, it rapidly shifts to generating the Jean-Paul Sartre content. This provides strong evidence that the Talker genuinely utilizes the semantic content encoded by the JEPA-Reasoner. Meanwhile, the fact that the generated output remains grammatically fluid despite this conflict proves that the Talker successfully handles the linguistic realization of the Reasoner's abstract concepts.

\section{K Value in Scaled Cosine Distance Loss}\label{app:k-val}
We tested $k$ from 1 to 6. All these $k$ values could produce a basic SST outcome that exhibits reasoning behaviors stated in all previous sections. With a careful tuning of $k$, we observed a stable improvement in the tree-search problem as shown in \cref{fig:k-value-comparison}:
\begin{figure}[ht]
    \centering
    \includegraphics[width=0.85\linewidth]{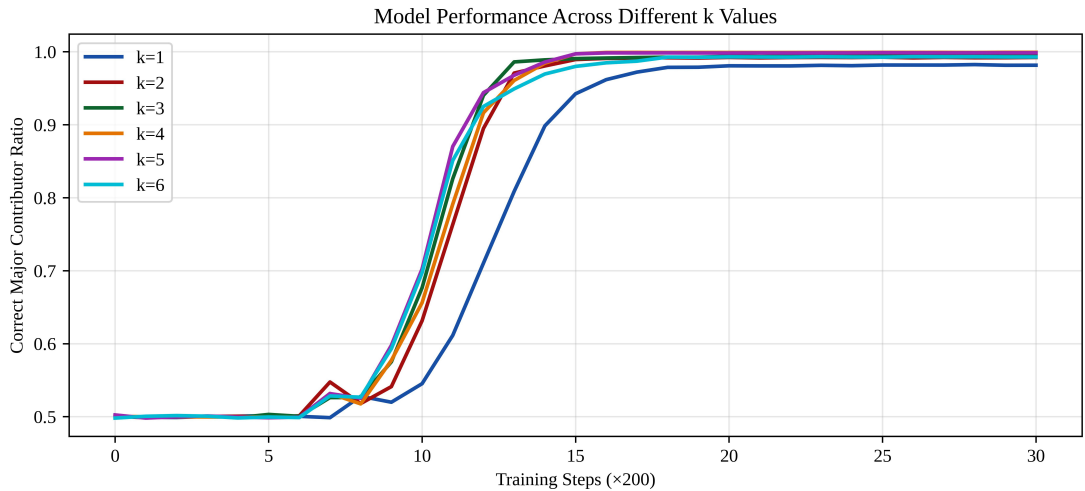}
    
    \caption{Changes of correct major contributor rate with training steps. Zoom in to see details.}
    \label{fig:k-value-comparison}
    
\end{figure}

We choose the correct major contributor (the correct next-step latent vector plays the most important role in current-step mixed latent vectors) rate as the metric since it directly relates to the correctness of future predictions. Considering that when $k=4$, the model gains the highest correct major contributor rate (zoom in to distinguish the line of $k=4$ from the line of $k=5$), we choose to continue our experiment with $k=4$.

\end{document}